\documentclass[runningheads]{llncs}
\usepackage{graphicx}

\usepackage{tikz}
\usepackage{comment}
\usepackage{amsmath,amssymb} 
\usepackage{color}

\usepackage{orcidlink}

\usepackage[accsupp]{axessibility}  

\usepackage[linesnumbered,ruled,vlined]{algorithm2e}

\usepackage{bm}
\usepackage{enumitem}
\newlist{inlinelist}{enumerate*}{1}
\setlist*[inlinelist,1]{%
      label=(\arabic*),
  }

\SetCommentSty{mycommfont}

\SetKwInput{KwInput}{Input}                
\SetKwInput{KwOutput}{Output}              

\newcommand{\x}{\bm{x}}
\newcommand{\y}{\bm{y}}
\newcommand{\btheta}{\bm{\theta}}
\newcommand{\wt}{\bm{w}}
\newcommand{\R}{\mathbb{R}}
\newcommand{\dataset}{\mathcal{D}}
\newcommand{\N}{\mathcal{N}}
\newcommand{\Jac}{\bm{J}}
\newcommand{\test}{{(t)}}
\newcommand{\DeltaKL}{\delta_\mathsf{KL}}
\newcommand{\oodmetric}{\mathrm{Unc}}
\newcommand{\E}{\mathbb{E}}
\DeclareMathOperator{\trace}{Tr}
\newcommand{\choice}{\bm{c}}
\def\sectionspace{\vspace{-0.5em}}
\def\subsectionspace{\vspace{-0.2em}}

\begin{document}
\pagestyle{headings}
\mainmatter
\def\ECCVSubNumber{183}  

\title{Data Lifecycle Management in Evolving Input Distributions for Learning-based Aerospace Applications} 

\titlerunning{Data Lifecycle Management for Learning-based Aerospace Applications}
%
\author{Somrita Banerjee\inst{1}\orcidlink{0000-0002-1340-8457} \and
Apoorva Sharma\inst{1} \and
Edward Schmerling\inst{1}\orcidlink{0000-0002-0552-9021} \and
Max Spolaor\inst{2} \and
Michael Nemerouf\inst{2} \and
Marco Pavone\inst{1}\orcidlink{0000-0002-0206-4337}}
\authorrunning{S. Banerjee et al.}
%
\institute{Stanford University, Stanford CA 94305, USA\\
\email{\{somrita,apoorva,schmrlng,pavone\}@stanford.edu} \and
The Aerospace Corporation, El Segundo CA 90245, USA\\
\email{\{max.spolaor,michael.k.nemerouf\}@aero.org}}
\maketitle

\addtolength{\belowcaptionskip}{-5mm}
\addtolength{\abovecaptionskip}{-2mm}

\vspace{-8mm}

\begin{abstract}
As input distributions evolve over a mission lifetime, maintaining performance of learning-based models becomes challenging. This paper presents a framework to incrementally retrain a model by selecting a subset of test inputs to label, which allows the model to adapt to changing input distributions. Algorithms within this framework are evaluated based on 
\begin{inlinelist}
\item model performance throughout mission lifetime and 
\item cumulative costs associated with labeling and model retraining.
\end{inlinelist}
We provide an open-source benchmark of a satellite pose estimation model trained on images of a satellite in space and deployed in novel scenarios (e.g., different backgrounds or misbehaving pixels), where algorithms are evaluated on their ability to maintain high performance by retraining on a subset of inputs.
We also propose a novel algorithm to select a \emph{diverse} subset of inputs for labeling, by characterizing the information gain from an input using Bayesian uncertainty quantification and choosing a subset that maximizes collective information gain using concepts from batch active learning. We show that our algorithm outperforms others on the benchmark, e.g., achieves comparable performance to an algorithm that labels 100\% of inputs, while only labeling 50\% of inputs, resulting in low costs and high performance over the mission lifetime.
\end{abstract}

\section{Introduction}
\paragraph{\textup{\textbf{Motivation.}}}
Learning-based models are being increasingly used in aerospace applications because of their ability to deal with high-dimensional observations and learn difficult-to-model functions from data. However, the performance of these models depends strongly on the similarity of the test input to the training data. During a mission lifetime, anomalous inputs and changing input distributions are very common. For example, a model that uses camera images to estimate the pose of a satellite could encounter out-of-distribution (OOD) inputs such as different backgrounds, lighting conditions, lens flares, or misbehaving pixels \cite{KisantalSharmaEtAl2020,MurrayBourlaiEtAl2021}. For rover dynamics models that are adaptive and learning-enabled, the OOD inputs could result from encountering novel terrain or sensor degradation \cite{WellhausenRanftlEtAl2020}. Without any intervention, these OOD inputs would then lead the model to produce incorrect predictions and potentially unsafe results. 

For safety-critical and high consequence applications, such as those in aerospace, 
it is crucial that the model remains trustworthy throughout the mission lifetime. 
In order to support long deployment lifetimes, during which the input distribution may evolve, it is critical to design models that are resilient to such changes. 
This highlights the need for frameworks like Aerospace's Framework for Trusted AI in High Consequence Environments \cite{SlingerlandMandrakeEtAl2022} and methods of retraining the model during the mission lifetime, so that it can adapt to evolving conditions. 
We refer to this challenge as the problem of ``data lifecycle management,'' entailing how to detect when a learning-based model may be uncertain or performing poorly, and how to design strategies to mitigate this poor performance, e.g., by referring to a human operator or fallback system, or obtaining labels for new inputs to retrain and adapt the model for continued high performance.

\begin{figure}[t]
\centering
\includegraphics[width=0.9\textwidth]{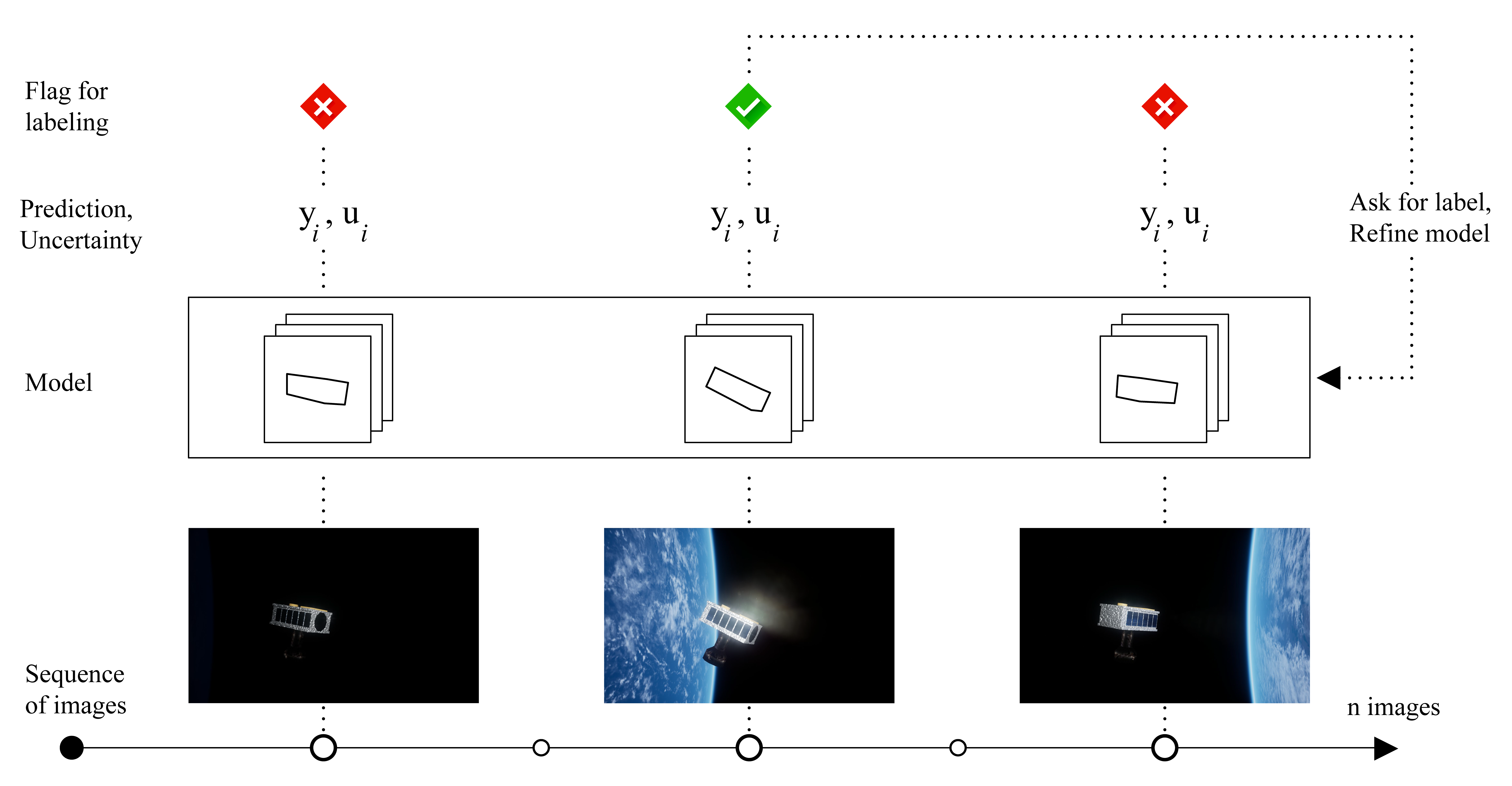}
\caption{Given a sequence of inputs, a model outputs a prediction and, potentially, an uncertainty. Data lifecycle management for a model requires adapting to changing input distributions, e.g., novel backgrounds, by periodically refining the model with inputs that were flagged, i.e., inputs that the model is uncertain about.}
\label{fig:datalifecycle_framework}
\end{figure}

Closing the data lifecycle loop is challenging because it requires storing data, downlinking data, obtaining oracle labels, and uplinking labels for retraining, all of which are expensive operations, and must usually be done in a batched episodic manner to ensure safety and computational feasibility. These challenges are discussed in more detail in Section \ref{sec:background-challenges}. In order to practically retrain models during their lifetimes, we need onboard real-time decision making about which inputs are most informative to store and label. In this work, we focus on managing the data lifecycle within the context of the framework shown in Figure \ref{fig:datalifecycle_framework}, which consists of the following steps: select a subset of test inputs for labeling, obtain labels for these inputs (either by expert humans or an oracle fallback system), and use these labels to retrain / adapt the model over time, so that the model can adapt to changing distributions. 

\paragraph{\textup{\textbf{Related work.}}}
The data lifecycle framework, as presented, can be broadly divided into two algorithmic components:
\begin{inlinelist}
    \item selecting a subset of inputs for labeling and
    \item retraining the model using these inputs. 
\end{inlinelist}
For the first task, techniques such as uncertainty quantification \cite{AbdarPourpanahEtAl2021} and OOD detection \cite{YangZhouEtAl2021} are useful for detecting inputs the model is most uncertain about, has not encountered previously, and is likely to perform poorly on. However, these techniques only provide information about a single test input and do not capture the mutual information correlation between multiple test inputs. A practical solution to selecting inputs requires augmenting these techniques with a notion of group uncertainty and choosing subsets that are collectively maximally informative. 
Choosing subsets of inputs that maximize information gain has been the focus of active learning \cite{Settles2012}, especially batch mode active learning \cite{HoiJinEtAl2006,GuoSchuurmans2007,SenerSavarese2018,PinslerGordonEtAl2019}. 
Some approaches for selecting inputs include labeling the inputs that lead to greatest expected error reduction \cite{RoyMcCallum2001}, that would change the model the most \cite{FreytagRodnerEtAl2014}, for which the model has the highest predictive uncertainty \cite{YangLoog2016}, or that maximize diversity and information gain \cite{PinslerGordonEtAl2019}. 
However, these works do not address how to quantify the information gain from an input. Our proposed algorithm bridges the gap by using Bayesian uncertainty to express the information gain from an input and techniques from batch active learning to choose a diverse subset.

The literature also lacks a benchmark to evaluate these algorithms in a full data lifecycle context. For example, the WILDS benchmark \cite{KohSagawaEtAl2021} is useful to evaluate the robustness of models to distribution shifts, but does not close the loop on evaluating the effect of labeling and retraining the model on the lifelong performance of the model. Similarly, the second algorithmic component, i.e., retraining the model, has been the focus of continual learning algorithms \cite{ParisiKemkerEtAl2019}, for which benchmarks such as CLEAR have been developed \cite{LinShiEtAl2021,HsuLiuEtAl2018} but these benchmarks do not consider the impact of judiciously selecting which inputs to label on the downstream retraining task. This paper addresses this gap by presenting a benchmark for the full data lifecycle.

\paragraph{\textup{\textbf{Contributions.}}}
As our first contribution, we propose a benchmark that evaluates algorithms for subselecting and retraining along two macro-level metrics
\begin{inlinelist}
\item model performance throughout mission lifetime and 
\item cumulative costs associated with labeling and retraining. 
\end{inlinelist}
While this benchmark is generalizable to any application, we provide open-source interfaces and test sequences from the ExoRomper dataset, developed by The Aerospace Corporation. 
The ExoRomper simulation dataset was assembled in support of Aerospace’s Slingshot program \cite{WeiherMabryEtAl2022}, which recently launched a commercial 12U smallsat bus, Slingshot 1, with 19 ``plug and play'' payloads, including ExoRomper. The ExoRomper payload is designed with thermal and visible light cameras facing a maneuverable model spacecraft and leverages an AI accelerator payload also on Slingshot 1 to experiment with machine learning-based pose estimation solutions. 
The ExoRomper dataset contains simulated camera images of the maneuverable model spacecraft as viewed on-orbit, along with ground truth labels of the satellite model’s pose. Test sequences constructed from this dataset include different types of OOD images which can be used to evaluate a pose estimation model and compare different algorithms in the context of a data lifecycle.  

In order to balance labeling costs and maintain model performance, we need onboard real-time decision making about which inputs are most informative to store and label. This leads us to our second contribution\textemdash a novel algorithm, Diverse Subsampling using Sketching Curvature for OOD Detection (DS-SCOD)\textemdash which selects a diverse subset of OOD inputs for labeling and retraining. This algorithm leverages Jacobians from a state-of-the-art OOD detection and uncertainty quantification algorithm, SCOD, \cite{SharmaAzizanEtAl2021} and concepts from Bayesian active learning \cite{PinslerGordonEtAl2019} to select a diverse subset of inputs for labeling, which are then used to adapt the model to changing distributions. We show that this algorithm outperforms other algorithms (naive, random, and uncertainty threshold-based) on the benchmark, i.e., results in high model performance throughout the mission lifetime, while also incurring low labeling costs.

\paragraph{\textup{\textbf{Organization.}}}
The paper is organized as follows. Section \ref{sec:background} discusses the challenges associated with data lifecycle management and introduces two tools that are instrumental to our approach: OOD detection algorithms and batch active learning methods. Section \ref{sec:prob_formulation_framework} describes the problem formulation for data lifecycle management. Section \ref{sec:benchmark} presents our open-source benchmark for evaluating data lifecycle algorithms. Section \ref{sec:DS-SCOD} introduces our novel data subselection algorithm, DS-SCOD. Section \ref{sec:results} presents results of evaluating our algorithm on the benchmark using the ExoRomper dataset where it outperforms other methods, e.g. achieves performance comparable to an algorithm that labels 100\% of inputs while only labeling 50\% of inputs. Finally, Section \ref{sec:conclusion} provides conclusions and future research directions.  

\section{Background} \label{sec:background}
\sectionspace
\subsection{Challenges for Data Lifecycle Management} \label{sec:background-challenges}
\subsectionspace
Data lifecycle management, especially for aerospace applications, has a number of challenges. We use the example of pose estimation for CubeSats \cite{PoghosyanGolkar2017}, such as the Slingshot satellite, to contextualize these challenges, which include:
\begin{enumerate}
\item Data collection and storage may be limited by onboard storage capacity, especially for large input sizes, constant streams of data, and having to store inputs for future labeling.
\item Data transmission may be limited by cost and practicality, e.g., CubeSats often have intermittent connectivity to ground stations, narrow downlink windows, and limited bandwidth \cite{PoghosyanGolkar2017}.
\item Operations associated with labeling are expensive, e.g., obtaining a label from a human expert.
\item Retraining the model is expensive and limited by onboard computation power \cite{DoranDaigavaneEtAl2022,PoghosyanGolkar2017}, especially for space hardware on CubeSats. Rather than retraining on each individual flagged input, it is more computationally feasible to retrain on batches of flagged inputs.
\item Lastly, building safe and trusted AI systems often requires model updates to be episodic and versioned in nature, in order to ensure adequate testing. This results in a batched, rather than streaming, source of ground truth labels as well as delays between flagging inputs and receiving labels.
\end{enumerate}

These challenges highlight the need for a judicious, real-time, and onboard decision-making algorithm that selects inputs for storage, downlinking, labeling, and retraining. In this work, we model some of these challenges by designing our algorithms to operate on batches and explicitly optimizing for low labeling cost. Next, we discuss two key tools that enable our algorithmic design.

\subsection{Out-of-Distribution Detection} \label{sec:background-scod}
\subsectionspace
A key requirement of data lifecycle management is to be able to detect when inputs to the model are no longer similar to those seen during training, i.e., detect when inputs are OOD and the model is likely to perform poorly. Detecting when inputs are OOD is useful in many ways, e.g., as a runtime monitor or anomaly/outlier detector so that the corresponding outputs can be discarded or a fallback system can be used.
OOD detection is also useful in deciding which inputs to use to adapt the model, so model performance does not degrade as input distributions evolve. A principled approach to OOD is to quantify the functional uncertainty of the model for each test input, i.e., how well does the training data determine the predictions at a new test input. This work employs the Bayesian approach to uncertainty quantification developed by Sharma et al., Sketching Curvature for Out-of-Distribution Detection (SCOD) \cite{SharmaAzizanEtAl2021}, which is summarized below.

Consider a pre-trained model $f$ with weights $\wt \in \R^N$ that maps from input $\x \in \mathcal{X}$ to the parameters $\btheta \in \Theta$ of a probability distribution $p_{\btheta}(y)$ on targets $\y \in \mathcal{Y}$, where $\btheta = f(\x, \wt)$. The model is trained on a dataset of examples $\dataset = \{\x_i,\y_i\}_{i=1}^M$.The prior over the weights of the model is $p(\wt)$ and the posterior distribution on these weights given the dataset is $p(\wt \mid \dataset)$. Characterizing the posterior and marginalizing over it produces the posterior predictive distribution 
\begin{equation}
    p(\y \mid \x) = \int p(\wt \mid \dataset) p_{\wt}(\y \mid \x) d\wt.
\end{equation}
\subsubsection{Estimating posterior offline.}
To tractably compute this posterior for Deep Neural Networks, SCOD uses the Laplace second order approximation of the log posterior $\log p(\wt \mid \dataset)$ about a point estimate $\wt^*$. If the prior on the weights is the Gaussian $p(\wt) = \N(0, \epsilon^2 I_N)$, the Laplace posterior is also a Gaussian, with mean equal to the point estimate $\wt^*$, and covariance given by 
\begin{equation}
\label{eqn:laplace_posterior}
\Sigma^* = \left(H_L + \epsilon^{-2} I_N \right)^{-1},
\end{equation}
where $H_L$ is the Hessian with respect to $\wt$ of the log likelihood of the training data, namely, $L(\wt) = \sum_{i=1}^M \log p_{\wt^*}(\y^{(i)} \mid \x^{(i)})$. The Hessian can in turn be well approximated by the dataset Fisher information matrix, i.e.,
\begin{equation}
\label{eqn:hessian}
    H_L = \frac{\partial^2 L}{\partial \wt^2} \biggr|_{\wt=\wt^*} \approx M F^\dataset_{\wt^*}.
\end{equation}
The dataset Fisher information matrix $F^\dataset_{\wt^*}$ describes the impact of perturbing the weights $\wt$ on the predicted output distribution for each output in the dataset $\dataset$, i.e.,
\begin{align}
    F^\dataset_{\wt^*} &= \frac{1}{M} \sum_{i=1}^M F^{(i)}_{\wt^*} \label{eqn:fisher_dataset} \\
     F_{\wt^*}^{(i)}(\x_i, \wt^*) &= \Jac_{f,\wt^*}^T 
     F_{\btheta}^{(i)}(\btheta)
     \Jac_{f,\wt^*} \label{eqn:fisher_indiv},
\end{align}
where $\Jac_{f,\wt^*}$ is the Jacobian matrix $\frac{\partial f}{\partial \wt}$ evaluated at $\left( \x_i, \wt^* \right)$ and $F_{\btheta}^{(i)}(\btheta)$ is the Fisher information matrix of the distribution $p_{\btheta}(y)$. In the case where $p_{\btheta}(y)$ is a Gaussian with mean $ \btheta$ and covariance $\Sigma$, the Fisher information matrix is given by 
\begin{equation}
\label{eqn:fisher_theta}
    F_{\btheta}^{(i)}(\btheta) ={\Sigma( \x^{\test}, \wt^* )}^{-1}.
\end{equation}
SCOD uses a matrix sketching-based approach to efficiently compute a low-rank approximation of the dataset Fisher $F^\dataset_{\wt^*}$, which in turn enables the computation of the Laplace posterior $\Sigma^*$ using Equations \ref{eqn:laplace_posterior}, \ref{eqn:hessian}, and \ref{eqn:fisher_dataset}. This is done offline.

\subsubsection{Online computation of predictive uncertainty.}
For a test input $\x^\test$, the predictive uncertainty is computed as the expected change in the output distribution (measured by a KL divergence $\DeltaKL$) when weights are perturbed according to the Laplace posterior $\Sigma^*$. Sharma et al. show that this expectation can be related back to the Fisher information matrix for this test input $F_{\wt^*}^\test$, i.e.,
\begin{align}
\label{eqn:uncertainty}
    \oodmetric(\x^\test) &= \underset{d\wt \sim \N(0, \Sigma^*)}{\E}\left[ \DeltaKL(\x^\test) \right] \nonumber \\
    &\approx \trace\left( F^\test_{\wt^*} \: \Sigma^* \right). 
\end{align}
Therefore, for a test input $\x^\test$, SCOD uses a forward pass to calculate the output $p(\y) = f(\x^\test,\wt)$, a backward pass to calculate the Jacobian $\Jac_{f,\wt^*}$ at $\x^{\test}$, and the output covariance $\Sigma\left( \x^{\test}, \wt^* \right)$. Then, using Equations \ref{eqn:fisher_indiv}, \ref{eqn:fisher_theta}, and \ref{eqn:uncertainty}, the predictive uncertainty $\oodmetric(\x^\test)$ can be efficiently computed. 
This predictive uncertainty can be thought of as a measure of epistemic uncertainty: if it is high, then the prediction for input $\x^{\test}$ is \textit{not} well supported by the training data, while if it is low, then the training data supports a confident prediction. Therefore, this estimate is useful for detecting OOD inputs.

\subsection{Bayesian Batch Active Learning} \label{sec:background-bal}
\subsectionspace
Detecting inputs that have high uncertainty allows those inputs to be flagged for labeling and retraining. However, in practice, a limited budget for retraining precludes selecting every uncertain input. 
While a naive solution might be to select the subset with the \emph{highest} uncertainties, this can lead to subsets with redundancy: including many similar inputs while leaving out other inputs that are also informative. Instead, it is important that we not select inputs greedily, but rather sets of \emph{diverse} inputs for which labels would jointly provide more information.
To do so, we take inspiration from active learning, where the goal is to select the most informative inputs, and, in particular, batch active learning, which selects a batch of informative inputs instead of a single input, allowing retraining to be done episodically rather than sequentially, further lowering labeling costs \cite{HoiJinEtAl2006,GuoSchuurmans2007,SenerSavarese2018,PinslerGordonEtAl2019}. This work closely utilizes the Bayesian batch active learning approach developed by Pinsler et al. \cite{PinslerGordonEtAl2019}, which is summarized below.

Consider a probabilistic predictive model $p(\y\mid \x, \btheta)$ parameterized by $\btheta \in \Theta$ mapping from inputs $\x \in \mathcal{X}$ to a probabilistic output $\y \in \mathcal{Y}$. The model is trained on the dataset $\dataset_0 = \{\x_i, \y_i\}_{i=1}^n$, resulting in the prior parameter distribution $p(\btheta | \dataset_0)$. During deployment, the model encounters a set of test inputs $\mathcal{X}_p = \{\x_i\}_{i=1}^m$. It is assumed that an oracle labeling mechanism exists which can provide labels $\mathcal{Y}_p = \{\y_i\}_{i=1}^m$ for the corresponding inputs. From a Bayesian perspective, it is optimal to label every point and use the full dataset $\dataset_p = \left( \mathcal{X}_p, \mathcal{Y}_p \right)$ to update the model. The resultant data posterior is given by Bayes' rule 
\begin{equation}
\label{eqn:posterior}
    p(\btheta | \dataset_0 \cup \left(\mathcal{X}_p, \mathcal{Y}_p\right)) = \frac{p(\btheta | \dataset_0) \, p( \mathcal{Y}_p | \mathcal{X}_p, \btheta)}{p(\mathcal{Y}_p | \mathcal{X}_p, \dataset_0)}.
\end{equation}
In practice, only a subset of points $\dataset' = (\mathcal{X}', \mathcal{Y}') \subseteq \dataset_p$ can be selected. From an information theoretic perspective, it is desired that the query points $\mathcal{X}'=\{\x_i\}_{i=1}^k$ (where $k \leq m$) are maximally informative, which is equivalent to minimizing the expected entropy, $\mathbb{H}$, of the posterior, i.e.,
\begin{equation}
    \label{eqn:information_theoretic_al}
    \begin{aligned}
    & \mathcal{X}^\ast = \underset{\mathcal{X}' \subseteq \mathcal{X}_p,\, | \mathcal{X}'| \leq k} {\text{arg min}}
    & & \mathbb{E}_{\mathcal{Y}' \sim p\left(\mathcal{Y}' | \mathcal{X}', \dataset_0\right)} \left[ \mathbb{H}\left[ \btheta | \dataset_0 \cup \left(\mathcal{X}', \mathcal{Y}'\right) \right] \right].  \\
    \end{aligned}
\end{equation}
Solving equation \ref{eqn:information_theoretic_al} requires considering all possible subsets of the pool set, making it computationally intractable. Instead, Pinsler et al. choose the batch $\dataset'$ such that the updated log posterior \mbox{$\log p(\btheta | \dataset_0 \cup \dataset')$} best approximates the complete data log posterior \mbox{$\log p(\btheta | \dataset_0 \cup \dataset_p)$}. The log posterior, following equation \ref{eqn:posterior} and taking the expectation over unknown labels $\mathcal{Y}_p$, is given by  
\small
\begin{equation}
\label{eqn:expected_log_posterior}
\begin{split}
\mathop\mathbb{E} \limits_{\mathcal{Y}_p} [\log p(\btheta | \dataset_0 \cup (\mathcal{X}_p, \mathcal{Y}_p))] &= \mathop\mathbb{E}\limits_{\mathcal{Y}_p} \left[\log p(\btheta | \dataset_0) + \log p(\mathcal{Y}_p | \mathcal{X}_p, \btheta) - \log p(\mathcal{Y}_p |\mathcal{X}_p, \dataset_0) \right] \\
&= \log p(\btheta | \dataset_0) + \sum_{i=1}^{m}  \Bigg( \underbrace{\mathop\mathbb{E} \limits_{\y_i} \left[ \log p(\y_i | \x_i, \btheta) \right] + \mathbb{H} \left[ \y_i | \x_i, \dataset_0 \right]}_{\text{$\mathcal{L}_i(\btheta)$}} \Bigg),
\end{split}
\end{equation}
\normalsize
where $\mathcal{L}_i(\btheta)$ is the belief update as a result of choosing input $\x_i \in \mathcal{X}_p$. The subset $\mathcal{X}'$ can now be chosen to best approximate the complete belief update, i.e., by choosing
\begin{align} 
\label{eqn:sparse_optimization}
\choice^\ast =& \underset{\choice} {\text{minimize}} \ \left\lVert\sum_i \mathcal{L}_i - \sum_i \choice_i \mathcal{L}_i \right\rVert^2 \nonumber\\
& \text{subject to}\quad c_i \in \{0, 1\} \quad \forall i, \nonumber\\
& \hspace{5em} \sum_i \choice_i \leq k.
\end{align}
Here $\choice \in \{0, 1\}^m$ is a weight vector indicating which of the $m$ points to include in the subset of maximum size $k$. Pinsler et al. show that a relaxation,
\begin{align} 
\label{eqn:relaxed_fw_optimization}
\choice^\ast =&\underset{\choice} {\text{minimize}} \ \left( \bm{1} - \choice \right)^T \bm{K} \left( \bm{1} - \choice  \right)\nonumber \\ 
&\text{subject to} \quad c_i \geq 0 \quad \forall i, \nonumber\\ 
& \hspace{5em} \sum_i c_i \left\lVert\mathcal{L}_i\right\rVert = \sum_i \left\lVert\mathcal{L}_i\right\rVert,
\end{align}
where $K$ is a kernel matrix $K_{mn} = \left< \mathcal{L}_m, \mathcal{L}_n \right>$, can be solved efficiently in real-time using Frank-Wolfe optimization, yielding the optimal weights $\choice^\ast$ after $k$ iterations. The number of non-zero entries in $\choice^\ast$ will be less than or equal to $k$, denoting which inputs (up to a maximum of $k$) are to be selected from the batch of $m$ inputs. Empirically, this property leads to smaller subselections as more data points are acquired. We leverage this Bayesian approach to selecting maximally informative subsets in our algorithm design, which enables maintaining high model performance while reducing labeling costs. In the next section, we discuss the framework and problem formulation guiding our approach.

\section{Problem Formulation and Framework}\label{sec:prob_formulation_framework}
\sectionspace
In this paper, we consider the problem of data lifecycle management as applied to model adaptation via retraining using the specific framework shown in Figure \ref{fig:datalifecycle_framework}. This framework assumes that the model receives inputs in a sequence of batches and that the model outputs a prediction for each input. Additionally, the model may also output an estimate of uncertainty for each input, such as the epistemic uncertainty estimate provided by SCOD. Based on the outputs for each batch, a decision must be made about which (if any) inputs to flag for labeling. If flagged, the input is labeled by an oracle and then used to retrain the model.

In the context of this framework, we seek to 
\begin{inlinelist}
    \item develop a benchmark that evaluates algorithms not just for their accuracy and sample efficiency on a single batch, but over lifecycle metrics such as average improvement in lifetime model performance and cumulative lifetime labeling cost, and
    \item develop an algorithm to select inputs for labeling, that are ideally diverse and informative, in order to minimize costs associated with labeling and retraining.
\end{inlinelist}
In support of the first goal, we propose an evaluation benchmark that is described in the next section. In support of the second goal, we present a novel algorithm for diverse subset selection in Section \ref{sec:DS-SCOD}.

\section{Evaluation Benchmark} \label{sec:benchmark}
\sectionspace
In this section, we present a benchmark to evaluate algorithms that decide which inputs to flag for labeling. A flagging algorithm is evaluated based on how well it manages the performance of a pre-trained model on a sequence of test input batches. The two metrics used for this evaluation are: the cumulative labeling cost, i.e., how many inputs were flagged for labeling, and the rolling model performance, i.e., the prediction loss of the model during the deployment lifetime. The ideal flagging algorithm keeps model performance high while incurring a low labeling cost. This benchmark procedure is described in Procedure \ref{alg:eval_benchmark}.
\vspace{-1.5em}
\begin{algorithm}[!ht] \label{alg:eval_benchmark}
\DontPrintSemicolon
  
  \KwInput{\textit{model}, \textit{test\_seq}, \textit{test\_labels}, \textit{flagger}}
  \KwOutput{\textit{labeling\_cost}, \textit{rolling\_performance}}
  Initialize \textit{flagged} vector and \textit{outputs} \;
  \ForEach{i, batch $\in$ test\_seq}
  {
    \textit{outputs}[\textit{i}] \textleftarrow \textit{model}(\textit{batch})    \tcp*{store model predictions} 
        \textit{flags} = \textit{flagger}(\textit{batch})  \tcp*{flag to request oracle labels}
        \If{any(flags)}
        {  
            \textit{flagged\_inputs} = \textit{batch}[\textit{flags}] \tcp*{store flagged inputs}
            Update \textit{flagged}[\textit{i}]\;
            \textit{true\_labels} = \textit{test\_labels}[\textit{i}][\textit{flags}]  \tcp*{get oracle labels}
            \textit{model} \textleftarrow refine\_model(\textit{flagged\_inputs}, \textit{true\_labels})\tcp*{refine model}
            
        }
        
  }
  \tcc{calculate benchmark metrics}
  \textit{labeling\_cost} = cumsum(\textit{flagged})\\
  \textit{rolling\_performance} = -[loss(\textit{outputs}[\textit{i}], \textit{test\_labels}[\textit{i}]) for \textit{i, batch} $\in$ \textit{test\_seq}]\\
  \Return{labeling\_cost, rolling\_performance}\;
\caption{Evaluation benchmark for ``flagging'' algorithms}
\end{algorithm}
\vspace{-1.75em}

We provide an open-source benchmark to test flagging algorithms\footnote{Code for the benchmark and our experiments is available at \url{https://github.com/StanfordASL/data-lifecycle-management}.}. This interface includes a pre-trained pose estimation model and test sequences drawn from the ExoRomper dataset, as well as a categorization of different ``types'' of out-of-distribution images, such as those with different backgrounds (Earth/space), lens flares, or with added sensor noise. These input sequences can be passed in sequentially to the model, or as batches. Additionally, the benchmark also contains implementations of some simple flagging algorithms to compare against, such as flagging every input (\texttt{naive\_true}), flagging none of the inputs (\texttt{naive\_false}), and randomly flagging $k$ inputs from each batch (\texttt{random\_k}). The benchmark also contains implementations of two proposed flagging algorithms: 
\begin{inlinelist}
    \item \texttt{scod\_k\_highest} (or \texttt{scod} for short), which wraps the pre-trained model with a SCOD wrapper that provides an estimate of epistemic uncertainty for each input. From each batch, the $k$ inputs with the highest SCOD uncertainties are flagged for labeling, and
    \item \texttt{diverse}, Diverse Subsampling using SCOD (DS-SCOD), which is a novel algorithm that uses the Jacobians from SCOD and Bayesian active learning to select a subset of inputs that are not only out-of-distribution but also \emph{diverse}. This algorithm is further described in the next section.
\end{inlinelist}

\section{Diverse Subsampling using SCOD (DS-SCOD)} \label{sec:DS-SCOD}
\begin{figure}[tb]
\centering
\includegraphics[width=0.95\textwidth]{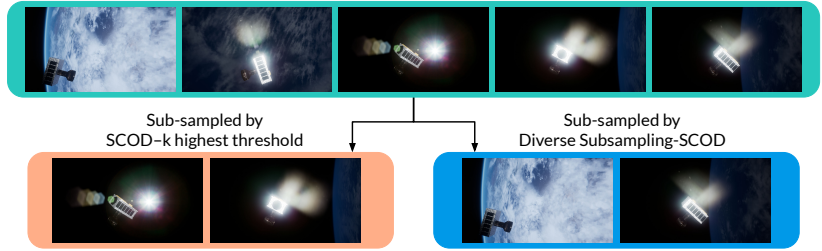}
\caption{Given a set of 5 random images drawn from the ExoRomper dataset and flagged as OOD by SCOD, a subset of 2 images is generated using SCOD-k highest threshold and Diverse Subsampling using SCOD. Although SCOD-k selects the 2 images with the highest predictive uncertainty, it appears, at least to a human eye, that the images selected by DS-SCOD are more semantically diverse.}
\label{fig:subselect-comparison}
\end{figure}

\sectionspace
To obtain performance improvement without high cost, it is necessary that the inputs chosen for labeling and retraining should provide information to the model that improves downstream performance, e.g., OOD inputs for which the model has high predictive uncertainty. However, an approach that greedily selects the $k$ most uncertain images from a batch may be suboptimal. Consider a batch of inputs containing multiple ``types'' of OOD, e.g., images with an Earth background or with a lens flare, where flagging the $k$ highest uncertainty images could lead to subsets containing only one type, while neglecting the information available from other types of OOD images. Downstream, this would lead to improved model performance for only the OOD type selected. In contrast, DS-SCOD seeks to capture the mutual correlation between multiple inputs, leading to a subset that contains diverse images and is maximally informative for the model. As discussed later in Section \ref{sec:results}, empirical testing shows that diverse subselecting leads to higher downstream performance, while allowing for low labeling cost. Figure \ref{fig:subselect-comparison} illustrates the different results from completing the task of subselecting 2 images from a set of 5 when using \texttt{scod} only, i.e., selecting the 2 images with the highest uncertainty, vs. using \texttt{diverse}, i.e., optimizing for diversity. 

DS-SCOD is based on two governing principles: 
\begin{inlinelist}
    \item that the inputs selected for labeling should provide information that improves downstream performance, e.g., inputs where the model has high uncertainty or OOD inputs and
    \item that the subselected inputs from each batch be chosen to maximize \emph{collective} information gain.
\end{inlinelist}
Therefore, the representation of information gain due to an input must lend itself to expressing the total information gain after selecting a subset of multiple inputs. An ideal choice for representing this information gain is the Bayesian posterior for each input that is provided by Bayesian OOD algorithms such as SCOD, as described in Section \ref{sec:background-scod}. Specifically, the Jacobian of the output with respect to the weights, $\Jac_{f,\wt^*}^{(i)}$, encodes the uncertainty for a single input, i.e., the impact of small weight changes on the output, and also encodes a correlation in uncertainty where $\Jac^{(i)}$ and $\Jac^{(j)}$ determine whether changes in output for inputs $i$ and $j$ are driven by the same set of weights. The key insight driving DS-SCOD is that OOD algorithms, which quantify uncertainty to detect differences between test inputs and a training distribution, are also naturally a powerful tool to detect differences between samples and quantify the uncertainty reduction resulting from selecting a batch for retraining.  

The core methodology employed by DS-SCOD is to use batch active learning to select the maximally informative batch, as described in Section \ref{sec:background-bal}, where the information value of an input is expressed using a Bayesian belief update that we obtain from SCOD, as described in Section \ref{sec:background-scod}. In DS-SCOD, we express the belief update, $\mathcal{L}_i(\btheta)$ as a result of choosing input $\x_i$, from Equation \ref{eqn:expected_log_posterior}, as
\begin{equation}
    \mathcal{L}_i(\btheta) =
    \left(F_{\btheta}^{(i)}(\btheta)\right)^{1/2}
    \: \Jac_{f,\wt^*}^{(i)}
\end{equation}
where $\Jac_{f,\wt^*}^{(i)}$ is the Jacobian of the output with respect to the weights evaluated at $\x_i$ and $F_{\btheta}^{(i)}(\btheta)$ is the Fisher information matrix of the output probability distribution $p_{\btheta}(y)$ from Equation \ref{eqn:fisher_theta}. Both of these expressions are calculated efficiently online by SCOD. Next, from each batch of size $m$, we select a maximum of $k$ inputs to flag for labeling by solving the optimization problem in Equation \ref{eqn:relaxed_fw_optimization}. Specifically, the optimization takes the form
\begin{align} 
\label{eqn:ds-scod}
\textbf{DS-SCOD:}& \nonumber\\
\choice^\ast = &\underset{\choice} {\text{minimize}} \ \left( \bm{1} - \choice \right)^T \bm{K} \left( \bm{1} - \choice  \right)\nonumber \\ 
&\text{where} \hspace{2.5em} K_{ij} = \left< \mathcal{L}_i, \mathcal{L}_j \right> = \trace\left( \left(F_{\btheta}^{(i)}\right)^{1/2} \: \Jac_{f,\wt^*}^{(i)} \: \Jac_{f,\wt^*}^{(j)T} \: \left(F_{\btheta}^{(j)}\right)^{1/2} \right) \nonumber\\
&\text{subject to} \quad c_i \geq 0 \quad \forall i, \nonumber\\ 
& \hspace{5em} \sum_i c_i \left\lVert \left(F_{\btheta}^{(i)}\right)^{1/2} \: \Jac_{f,\wt^*}^{(i)} \right\rVert = \sum_i \left\lVert \left(F_{\btheta}^{(i)}\right)^{1/2} \: \Jac_{f,\wt^*}^{(i)} \right\rVert.
\end{align}
This optimization is solved using the Frank-Wolfe algorithm in real-time. By minimizing the difference between the belief update after choosing all $m$ points and after choosing inputs according to $\choice$, the loss of information is minimized. This results in a maximally informative subsample of inputs, without exceeding the given labeling budget $k$. In the next section, we discuss the results of testing the DS-SCOD algorithm on the benchmark.

\section{Experimental Results}\label{sec:results}
\begin{figure}[bt]
\centering
\includegraphics[width=0.99\textwidth]{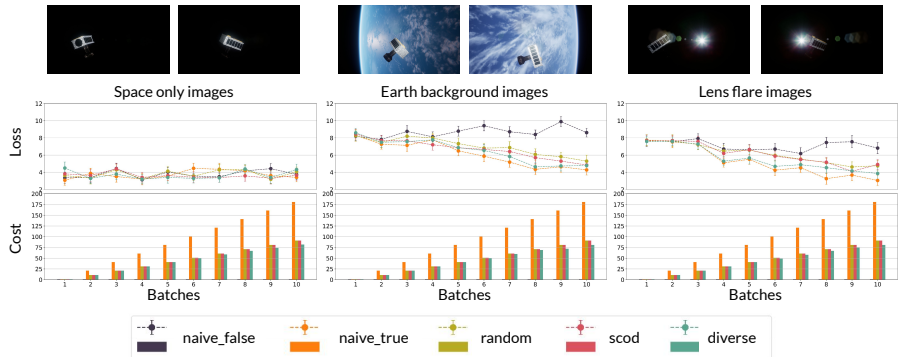}
\caption{Comparing algorithm performance on images of a single category. For each successive batch, the average loss of the model and the cumulative cost are plotted for each of the algorithms: naive\_false, naive\_true, random, scod, and diverse. For the latter three algorithms, the relabeling budget is set to 50\%. }
\label{fig:categories}
\end{figure}

\sectionspace
The benchmark described in Section \ref{sec:benchmark} was used to test DS-SCOD and other data lifecycle algorithms. The task for the benchmark was to estimate pose of a satellite given a camera image. The ExoRomper dataset was filtered into three categories: images with a space background, with Earth in the background, and with a lens flare (examples in Figure \ref{fig:categories}). A pose estimation model was trained on 300 images drawn \emph{only} from the space background set (considered ``in-distribution'') using a deep CNN architecture inspired by \cite{ChenCaoEtAl2019}, i.e., the winner of the Kelvins satellite pose estimation challenge organized by ESA and Stanford University \cite{KisantalSharmaEtAl2020}. The benchmark supports constructing test sequences by randomly sampling images from a configurable mix of categories. Our evaluation test sequence consisted of 100 batches of 20 images sampled from all three categories, with increasing numbers of degraded pixels ($\leq 10\%$), to simulate degrading sensors. From each batch, inputs could be flagged by the subselection algorithm. If any inputs were flagged, the corresponding ground truth labels were looked up, and the model was retrained on those labels for 200 epochs at a low learning rate of 0.001, before the start of the next batch. Thus, the evaluation was episodic but without delays between batch flagging and retraining. 

\begin{figure}[tb]
\centering
\includegraphics[width=0.85\textwidth]{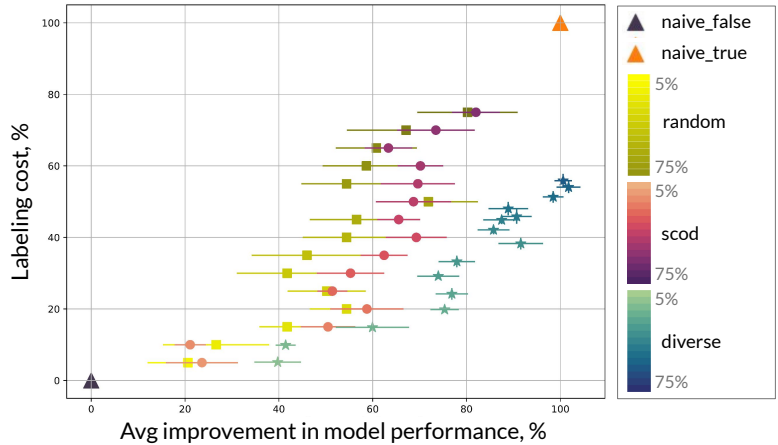}
\caption{Comparing performance of the naive, random, scod, and diverse (DS-SCOD) algorithms in terms of average improvement in model performance over the lifecycle and the labeling cost. The latter three algorithms were evaluated with labeling budgets between 5\% and 75\% at intervals of 5\%. The metrics are reported in percentages relative to the naive\_false (0\%) and naive\_true (100\%) algorithms.}
\label{fig:benchmark}
\end{figure}

As shown in Figure \ref{fig:benchmark}, the algorithms were evaluated on two metrics: the improvement in model performance as a result of retraining, averaged over the full test sequence, and the labeling cost, which is the number of images flagged for labeling expressed as a percentage of the total number of images. For some algorithms, we set a value k that determines what percentage of inputs the algorithm is allowed to select for labeling, henceforth referred to as the \textit{budget}. The following algorithms were tested:
\begin{itemize}
    \item \texttt{naive\_false}: This simple algorithm never flags any image. Thus, the labeling cost is zero. The performance of this algorithm is used as a baseline for other algorithms, therefore the percentage improvement in model performance is also zero.
    \item \texttt{naive\_true}: This simple algorithm flags every image. Thus, the labeling cost is 100\%. The performance of this algorithm improves due to the retraining. Since no algorithm can do better than having every input labeled (barring stochasticity, as we discuss later), we use the performance of this algorithm as the 100\% benchmark. 
    \item \texttt{random k\%}: This algorithm randomly selects k\% of inputs from each batch for labeling. This budget is varied between 5\% and 75\%, at intervals of 5\%. 
    \item \texttt{scod k\%}: This algorithm wraps the pre-trained model with a SCOD wrapper which computes the uncertainty for each input. From each batch, the top k\% of inputs with the highest SCOD uncertainties are flagged for labeling. 
    \item \texttt{diverse k\%}: This algorithm uses DS-SCOD, as presented in Section \ref{sec:DS-SCOD}, to select not only uncertain inputs, but also a diverse subset of inputs. For each batch, the optimization problem from Equation \ref{eqn:ds-scod} is solved efficiently. This results in flagging \emph{up to} k\% of inputs, in contrast to the previous algorithms that select a fixed, i.e. exactly k\%, number of inputs. 
\end{itemize}

We observe from Figure \ref{fig:benchmark} that random sampling leads to an improvement in performance; periodically retraining the model allows it to adapt to distribution shifts.
SCOD-k performs better than random sampling, on average, for a given budget since it picks the most uncertain inputs which results in higher information gain and downstream model performance. 
Diverse subsampling, DS-SCOD, outperforms random and SCOD-k by achieving better performance for a given budget. We observe that when the labeling budget is low, e.g., 10\% to 25\%, DS-SCOD expends the full budget, i.e., chooses exactly k\% of inputs to label. However, due to the relaxed optimization process, when given a higher budget, e.g., 40\% to 75\%, DS-SCOD uses only a fraction of the budget, not exceeding 60\% in Figure \ref{fig:benchmark}. Remarkably, DS-SCOD achieves performance comparable to \texttt{naive\_true} while incurring only 50\% of the labeling cost. DS-SCOD's performance sometimes slightly exceeds 100\% on our benchmark, i.e., makes slightly better predictions than \texttt{naive\_true}, due to the stochastic nature of the stochastic gradient descent we use to retrain the model. Overall, DS-SCOD results in the highest average lifetime model performance, while reducing labeling cost.

Additionally, the algorithms were benchmarked on test sequences drawn only from \emph{one} of the three categories, as shown in Figure \ref{fig:categories}. The labeling budgets were set to 50\%. For the space-only images, all the algorithms resulted in relatively constant test loss over time. This result is unsurprising because the model was trained on space background images, and likely reached convergence on these ``in-distribution'' images. For both the Earth background and lens flare images, all of the labeling algorithms, naive\_true, random, scod, and diverse, led to a decrease in loss over time, indicating that retraining is effective in increasing model performance when encountering OOD inputs. We observe that naive\_true achieves the greatest reduction in loss by labeling every input in each batch. Further, we note that the diverse (DS-SCOD) algorithm achieves a reduction in loss most comparable to naive\_true while accruing a labeling cost of less than 50\% that of naive\_true. Therefore, we show that DS-SCOD offers a cost-effective solution to adapting learning-based models to changing input distributions.

\section{Conclusion} \label{sec:conclusion}
\sectionspace
In this work, we discussed the challenges associated with data lifecycle management for learning-based components. We presented a framework for online adaptation by labeling a subset of test inputs and using the labels to retrain the model. We provided a benchmark for evaluating algorithms, that choose which subset of test inputs should be labeled, in the context of two macro-level metrics: the average model performance over the lifecycle and the cost of labeling. Lastly, we presented a novel subsampling algorithm, Diverse Subsampling using SCOD (DS-SCOD), inspired by ideas from Bayesian uncertainty quantification and Bayesian batch active learning. We showed that this algorithm, when evaluated on the benchmark, achieves performance comparable to labeling every input while incurring 50\% of the labeling costs. 

Interesting future directions include adding a delay between flagging inputs for labels and receiving labels for retraining, to better simulate the data lifecycle \cite{GrzendaGomesEtAl2020}. Another direction would be to expand the benchmark to other applications beyond satellite pose estimation, as well as different tasks and types of OOD inputs, since algorithm performance may be better suited to some tasks than others. Another promising research direction is to evaluate the effect of retraining algorithms on the improvement in model performance and design a retraining algorithm that makes maximal use of the information gain from diverse labeled inputs while avoiding catastrophic forgetting, e.g., by borrowing concepts from transfer learning and continual learning \cite{WeissKhoshgoftaarEtAl2016,ParisiKemkerEtAl2019}. 

\paragraph{\textup{\textbf{Acknowledgments.}}} This work is supported in part by The Aerospace Corporation's University Partnership Program, by DARPA under the Assured Autonomy program, and by the Stanford Graduate Fellowship (SGF). The NASA University Leadership initiative (grant \#80NSSC20M0163) provided funds to assist the authors with their research, but this article solely reflects the opinions and conclusions of its authors and not any NASA entity. The authors would like to thank Rohan Sinha and the reviewers for their helpful comments.
%
%
\bibliographystyle{splncs04}
\bibliography{main,ASL_papers}
\end{document}